\documentclass[letterpaper, 10 pt, conference]{ieeeconf}  % Comment this line out if you need a4paper
\IEEEoverridecommandlockouts

\usepackage{cite}
\usepackage{amsmath,amssymb,amsfonts}
\usepackage{algorithmic}
\usepackage{multirow} 
\usepackage{graphicx}
\usepackage{textcomp}
\usepackage{xcolor}
\usepackage{colortbl}  
\usepackage{booktabs}
\usepackage[colorlinks=true, linkcolor=cyan, urlcolor=magenta]{hyperref}
\usepackage{caption} 
\usepackage{makecell}
\usepackage{tabularx}
\usepackage{cuted} % <--- 加這個！

\def\BibTeX{{\rm B\kern-.05em{\sc i\kern-.025em b}\kern-.08em
    T\kern-.1667em\lower.7ex\hbox{E}\kern-.125emX}}

\newcolumntype{C}[1]{>{\centering\arraybackslash}p{#1}}
\newcolumntype{M}[1]{>{\centering\arraybackslash}m{#1}}

\begin{document}

% === Title & Authors ===
\title{Controllable Collision Scenario Generation via Collision Pattern Prediction}

% \author{Anonymous}
\author{
    Pin-Lun Chen$^{1}$, Chi-Hsi Kung$^{2}$, Che-Han Chang$^{3}$, Wei-Chen Chiu$^{1}$, Yi-Ting Chen$^{1,\dagger}$
    \thanks{ † Corresponding Author.}%
    \thanks{$^{1}$ Department of Computer Science, National Yang Ming Chiao Tung University, Hsinchu City, Taiwan. {\tt\small yp201141413.en11, ychen@nycu.edu.tw}}%
    \thanks{$^{2}$ School of Informatics, Computing, and Engineering, Indiana University, Bloomington, Indiana, USA. {\tt\small kung@iu.edu}}%
    \thanks{$^{3}$ MediaTek Inc.}
}

% \date{Preprint under review at ICRA 2026}

% % === Teaser + Title ===
\maketitle
% % \noindent\vskip -20mm
% \begin{figure}[t]

% % \begin{strip}

%   \centering
  
%   % \vspace{-25mm}
%   % \includegraphics[width=\textwidth]{Figures/Teaser_v5.pdf}
%   \includegraphics[width=\columnwidth]{Figures/Teaser_v5.pdf}
  
%   \captionof{figure}{Autonomous vehicles require diverse and structured collision scenarios to support comprehensive safety validation. Our proposed framework enables users to specify the desired collision type and time-to-accident (TTA) to generate attacker trajectories that realize controllable collision scenarios for autonomous vehicle (AV) testing. 
% %We further utilize generated scenarios to evaluate and improve motion planners.
% \end{figure}
% % \end{strip}

% === Abstract ===
\begin{abstract}

Evaluating the safety of autonomous vehicles (AVs) requires diverse, safety-critical scenarios, with collisions being especially important yet rare and unsafe to collect in the real world.
Therefore, the community has been focusing on generating safety-critical scenarios in simulation. 
However, controlling attributes such as collision type and time-to-accident (TTA) 
% in simulation 
remains challenging. 
% has been highlighted as a key challenge for systematic evaluation, but existing methods lack scalability or converge to limited outcomes.
%
We introduce a new task called controllable collision scenario generation, where the goal is to produce trajectories that realize a user-specified collision type and TTA, to investigate the feasibility of automatically generating desired collision scenarios. 
To support this task, we present COLLIDE, a large-scale collision scenario dataset constructed by transforming real-world driving logs into diverse collisions, balanced across five representative collision types and different TTA intervals.
We propose a framework that predicts \textit{Collision Pattern}, a compact and interpretable representation that captures the spatial configuration of the ego and the adversarial vehicles at impact, before rolling out full adversarial trajectories. 
%
% This strategy enables and scalable control of collision scenarios.
%
Experiments show that our approach outperforms strong baselines in both collision rate and controllability. Furthermore, generated scenarios consistently induce higher planner failure rates, revealing limitations of existing planners.
We demonstrate that these scenarios fine-tune planners for robustness improvements, contributing to safer AV deployment in different collision scenarios.
Additional generated scenarios are available at this  \href{https://plchen86157.github.io/conditional_scenario_generation/}{project webpage}.

\end{abstract}

% === Teaser + Title ===
% \maketitle
% \noindent\vskip -20mm

% \end{strip}
% --- Content ---
\section{Introduction}

Evaluating the reliability of autonomous vehicles (AVs) demands testing in diverse, safety-critical scenarios with varying attributes, including scenario categories (e.g., lane change, junction crossing), traffic participant types and states (e.g., time to accident), and road topologies~\cite{ding2023survey, schutt20231001, gao2025foundation, wang2024survey,Kung_2024_CVPR}. 
Among these, collision scenarios stand out as especially critical, since they directly test an AV’s ability to anticipate, react, and ensure safety under high-risk conditions.
However, collision scenarios are statistically rare and inherently unsafe for testing in the real world.
Therefore, the community has been focusing on generating safety-critical scenarios in simulation~\cite{ding2023survey, liu2024curse} and controlling attributes for diverse scenario generation~\cite{ding2023survey}. 

In this work, we propose a new task: controllable collision scenario generation. Given a target collision type and a desired time-to-accident (TTA), the goal is to generate trajectories that realize the specified collision. 
Prior work has primarily focused on adversarial trajectories that force a crash with the test vehicle.
However, they lack explicit controllability over key aspects of the scenario, such as varying the granularity of collision types or adjusting the time-to-accident. 
For example, recent advances~\cite{rempe2022generating,hanselmann2022king} can generate accident-prone scenarios, but these are often limited to specific accident types, such as rear-end or head-on crashes (see Fig. 6 in~\cite{rempe2022generating}). 
More recently, conditional generation with diffusion models~\cite{rempe2023trace, gu2022stochastic} has opened new possibilities for controllable collision scenario generation, yet existing formulations have not addressed fine-grained controllability. 
These gaps motivate our investigation into what it would take to achieve controllable collision scenario generation.
%
%
% that eventually result in a crash, our formulation emphasizes explicit control over the collision outcome. 
%
% One traditional scenario generation approaches offer fine-grained control by manually predefining collision configurations, agents, and TTA, but they require extensive manual effort to program agent behaviors, especially under
% diverse scene topologies and interactions, which limit scalability and generalization. 
%
% Adversarial-based methods~\cite{ding2020learning, klischat2019generating, rempe2022generating} automate generation by perturbing trajectories toward collisions, yet their optimization pipelines focus narrowly on collision occurrence and lack mechanisms to specify collision type or TTA, often limited to certain types such as rear-end or head-on crashes~\cite{rempe2022generating,hanselmann2022king}. Recent diffusion-based predictors have opened up new possibilities for conditional generation~\cite{xie2024advdiffuser, xu2025diffscene}, but the guidance can't enforce such precise spatial relations or timing.

\begin{figure}[t!]

% \begin{strip}

  \centering
  
  % \vspace{-25mm}
  % \includegraphics[width=\textwidth]{Figures/Teaser_v6.pdf}
  \includegraphics[width=\columnwidth]{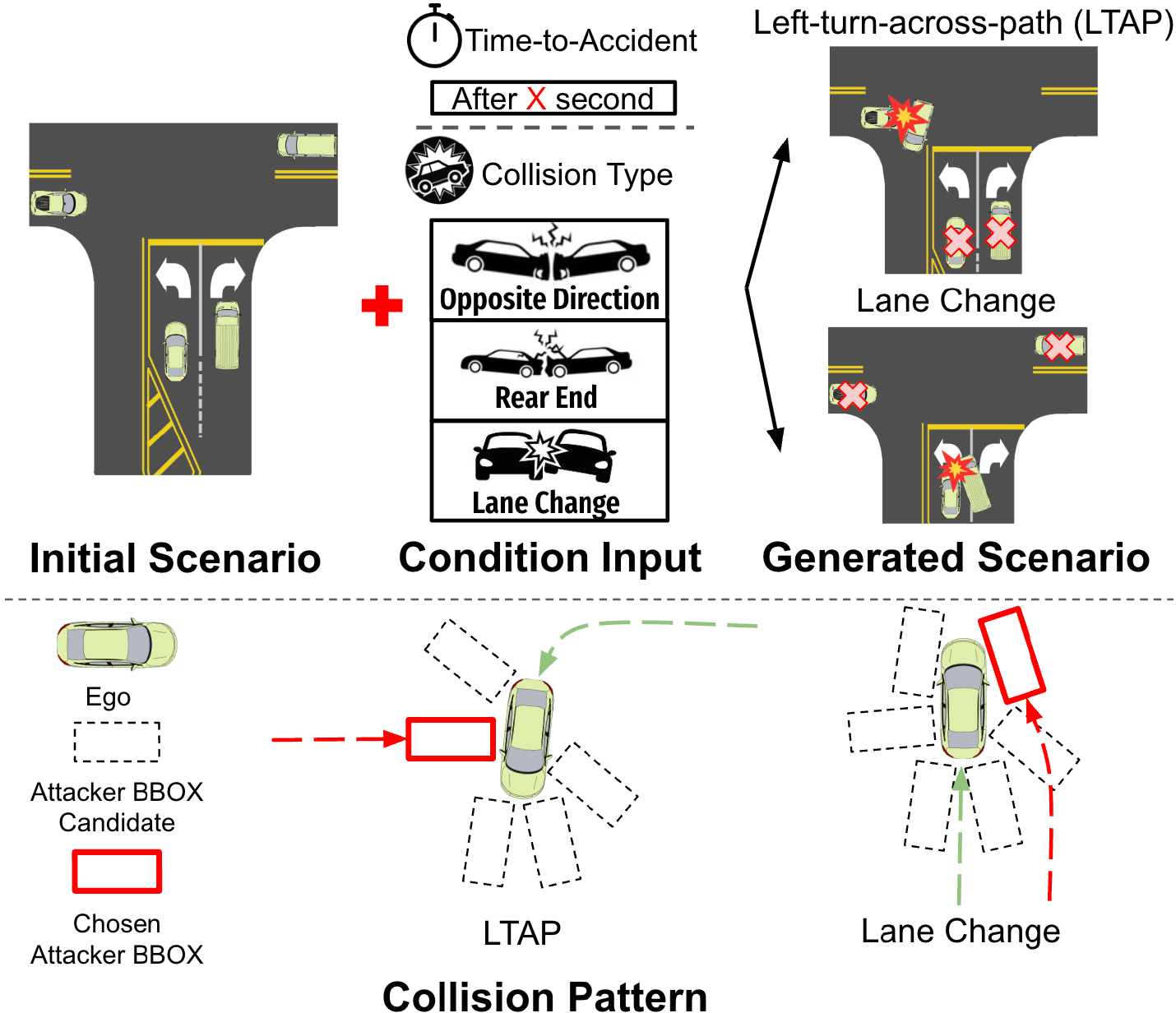}
  
  \captionof{figure}{
  Given an \textit{initial scenario}, users specify condition inputs including 
collision type and time-to-accident (TTA). The framework predicts a 
\textit{Collision Pattern}, defined as the relative configuration between the 
ego vehicle (green car) and the attacker (red bounding box) at the collision moment. 
This predicted pattern then guides the trajectory planner to generate a 
feasible attacker motion (red arrow), resulting in controllable \textit{generated scenarios} 
such as lane change or LTAP.
  % % Autonomous vehicles require diverse and structured collision scenarios to support comprehensive safety validation. 
  % Our proposed framework enables users to specify the desired collision type and time-to-accident (TTA) to generate attacker trajectories that realize controllable collision scenarios.
  % % for autonomous vehicle testing. 
  % The framework centers on the concept of a \textit{Collision Pattern} composed of the relative position,  heading, and roles of ego and adversarial vehicles at the moment of collisions. 
  % \textcolor{red}{unclear how the collision pattern is used for scenario generation}
  % % Such controllability is crucial for systematically evaluating how AV planners respond to diverse hazardous conditions.
%
% \ychen{why controllable collision scenarios
% for autonomous vehicle testing is a big matter!!}  
%We further utilize generated scenarios to evaluate and improve motion planners.
}
  \label{fig:teaser}
\end{figure}

To this end, we introduce COLLIDE, a large-scale collision scenario dataset specifically designed for controllable collision scenario generation. 
Existing collision scenario datasets~\cite{ chan2017anticipating,kataoka2018drive,suzuki2018anticipating,kim2019crash,you2020traffic,xiao2024hazardvlm,kung2024riskbench} lack collision-type annotations and do not provide scenarios with varying time-to-accident (TTA).
%
% thereby limiting their capability to support the study of controllable safety-critical scenario generation.
%
% Using a novel pipeline, we transform non-collision trajectories from real-world driving logs into diverse collision scenarios. 
To address this, we propose an automatic pipeline that transforms non-collision trajectories from real-world driving logs into diverse collision scenarios.
Specifically, we select target collision types based on common accident categories defined by the U.S. National Highway Traffic Safety Administration (NHTSA)~\cite{najm2013description}.
For a given ego trajectory, we choose a timestamp and extract the corresponding ego bounding box as a reference. An adversarial vehicle is then placed at all feasible spatial locations. 
The configuration that satisfies the definition of the desired collision type (see the bottom of Fig.~\ref{fig:teaser}) is selected.
%
% We refer to this configuration as the .”
%
Finally, given the spatial configuration of the ego and adversarial vehicles and the specified TTA, we use a quintic polynomial planner to generate a plausible trajectory.
This pipeline enables scalable and controllable collision scenario generation, supporting systematic evaluation of the corresponding algorithms.
This raises a question: how can we automatically generate collision scenarios given a collision type and TTA?
We argue that the key to generating collision scenarios lies in contextualizing \textit{collision type}.
For example, a \textit{Lane Change} collision evokes a rough pattern, as illustrated in the bottom part of Fig.~\ref{fig:teaser}. 
A Left-Turn-Across-Path (LTAP) collision would be a different pattern.
% that guides the generation of plausible trajectories.
%
Once we have a rough pattern, we can generate the corresponding plausible trajectories.
We term these patterns as \textit{Collision Pattern}, which encodes the relative position, heading, and roles of the ego and adversarial vehicles at the moment of impact. 
Rather than modeling full trajectories, \textit{Collision Patterns} focuses on the critical endpoint and offers a compact and interpretable representation for collision scenario generation.
%
% Another example a rear-end on a straight road and an LTAP at an intersection exhibit distinct spatial and temporal patterns when a collision happens. 
%
% By treating the collision pattern as an intermediate target, the model can plan trajectories that realize user-specified outcomes without error accumulation from stepwise prediction.
% that directly aligns with user-specified intent.
%This enables the model to plan trajectories that fulfill the user-specified intent.

Our framework is inspired by region proposal networks~\cite{ren2016faster} in object detection, where a rough location is first estimated and then refined. 
Similarly, we first predict a coarse collision pattern as an anchor and refine it to determine the precise spatial configuration. Based on this configuration, we generate the full adversarial trajectory. This 
outcome-driven and coarse-to-fine strategy enables effective generation, avoiding error accumulation from stepwise prediction while improving controllability and flexibility.

We conduct experiments on the COLLIDE dataset and benchmark conditional trajectory prediction baselines, including SGAN~\cite{gupta2018social}, MID~\cite{gu2022stochastic}, and STRIVE~\cite{rempe2022generating}. 
Our method consistently generates collisions that match user-specified time-to-accident (TTA) and collision types, achieving higher collision rates and thus more plausible scenarios. 
We further evaluate the generated scenarios on three rule-based motion planners—IDM~\cite{treiber2000congested}, a rule-based planner~\cite{rempe2022generating}, and PDM~\cite{dauner2023parting}—and find that our method induces more planner failures than C-STRIVE, exposing critical blind spots in existing planning strategies. 
% Finally, we demonstrate that these scenarios can tune hyperparameters of PDM, enabling it to better handle complex interactions such as missed yields or junction collisions.
To further evaluate the impact of the generated scenarios, we use them to update AV planners and found that ours can offer
%significant 
improvement in terms of collision rate in diverse dangerous scenarios.

Our contributions are summarized as follows:
\begin{itemize}
    \item We propose controllable collision scenario generation, a new task designed for comprehensive and systematic evaluation of autonomous vehicle safety. 
    
    \item We introduce an automatic data collection pipeline that accounts for collision types and TTA, enabling benchmarking of controllable collision scenario generation algorithms.  

    \item We introduce a compact and interpretable representation \textit{Collision Pattern}, which encodes the relative position, heading, and roles of the ego and adversarial vehicles at the moment of impact. 
    % \item A novel scenario generation model designed for controllability by focusing on the \textit{collision pattern}.

    \item We demonstrate that our framework can effectively generate user-specified scenarios, outperforming strong baselines. Moreover, the generated scenarios reveal limitations in rule-based motion planners and can be used to improve their robustness.  
    
    % theincluding the  showing our model's state-of-the-art performance in terms of both collision rate and controllability.
    
    % \item Downstream experiments that assess the effectiveness of our generated scenarios on 
    % rule-based 
    % planners and demonstrate that our model can improve the robustness of planners.
    %planners' ability.
\end{itemize}

\section{Related Work}
\label{sec:citations}

%======================Architecture=============================
%     {\noindent \bf 
% 1. Safety-critical driving scenario generation. \\ Briefly introduce the history of safety-critical driving scenario generation. However, the current safety-critical scenario generation (SCSG) methods fail to conditionally generate specific and diverse types of scenarios. For example, King and STRIVE.  This limitation can lead to inefficient AD testing. \\
% Furthermore, condition scenario generation is rising, diffusion model can detailed 
% 2. Traffic simulation \\ (we need to mention our multi-agent trajectory prediction baselines here.)
% Generative model can simulate traffic flows, and some SCSG methods use traffic simulation to generate safety-critical scenarios. However, adding high-level condition is hard.
% }
%======================Architecture=============================

{\noindent \bf Scenario Generation.}  
Existing approaches can be categorized as \textit{data-driven}, \textit{adversarial}, and \textit{knowledge-based}, according to~\cite{ding2023survey}.  
Data-driven methods~\cite{knies2020data, scanlon2021waymo, ding2018new} extract rare collision events from large-scale real-world datasets \cite{krajewski2018highd}, but such events are sparse and highly imbalanced in terms of both collision type and urgency (TTA), limiting their value for training or evaluation.  
Adversarial methods~\cite{rempe2022generating, wang2021advsim, hanselmann2022king}  perturb agent trajectories within learned traffic dynamics models~\cite{suo2021trafficsim} to induce failures, but they often bias toward specific failure modes and lack explicit controllability over collision semantics.
% Adversarial methods~\cite{rempe2022generating, wang2021advsim, hanselmann2022king} generate collisions by perturbing agent trajectories within a learned traffic dynamics model, such as TrafficSim~\cite{suo2021trafficsim}, which predicts realistic multi-agent motions 
% from large-scale trajectory datasets. These perturbations are typically applied by altering the velocity, heading, or acceleration of an adversarial vehicle in a way that drives the ego-vehicle into conflict.
% %perturb agent trajectories based on learned traffic dynamics to induce collisions.
% While adversarial methods are effective at generating failure cases, they tend to converge on a narrow set of more likely outcomes, such as rear-end or head-on collisions.
% %, due to the lack of explicit control over how and when the crash should happen. 
% This bias arises because their perturbation pipelines are typically regularized by learned traffic priors, which favor straight-line following or opposing flows. 
% % This bias arises because common traffic priors regularize perturbation pipelines, which favor straight-line following and opposing flows. 
% In addition, their optimization objective seeks minimal trajectory changes for ensuring realism and optimization stability, which naturally produces simpler collision geometries while sacrificing controllability over rarer cases such as junction intrusions or lane change crashes.
Knowledge-based approaches~\cite{ding2023causalaf, huang2024cadre, kung2024riskbench, xu2022safebench} incorporate rule-based priors or pre-defined templates to guarantee coverage of certain collision types but are not scalable, as increasing intra-class variation requires substantial manual effort and does not generalize efficiently.
% Knowledge-based approaches~\cite{ding2023causalaf, huang2024cadre, kung2024riskbench, xu2022safebench} incorporate rule-based priors or semantic graphs to constrain the generation process.
% For instance, SafeBench~\cite{xu2022safebench} and RiskBench~\cite{kung2024riskbench} rely on pre-defined templates, which guarantee coverage of certain collision types but are not scalable, as increasing intra-class variation requires substantial manual effort and does not generalize efficiently. 
% Similarly, CausalAF~\cite{ding2023causalaf} is confined by a hand-crafted causal graph, preventing diverse variations or generalization to other scenarios.

Our formulation does not fall into the existing three categories. Instead, it bridges data-driven realism by learning the distribution from COLLIDE and knowledge-based structure via collision pattern supervision. %, enabling controllability that adversarial optimization methods fail to achieve.
% To explicitly control collision type and TTA in a scalable manner, we frame the task as a \textit{collision-aware trajectory prediction problem}. First, the model predicts a collision pattern, a structured representation of the relative spatial configuration between the ego and the adversarial vehicle at the moment of impact. Conditioned on this intermediate target, the adversarial vehicle’s trajectory is then rolled out to realize the specified outcome.
%: given a structured interaction endpoint, collision pattern, our model generates the attacker's trajectory that fulfills the desired collision type and time-to-accident (TTA) constraint. 
This formulation directly addresses the limitations of prior work: by training on synthesized collision trajectories derived from real-world logs
%and grounded in map context
, we mitigate the data scarcity faced by real-world–only methods; by conditioning on structured collision patterns and TTA, we enable intra-category diversity that knowledge-based methods, which rely on rule-based or heuristic-driven methods, often fail to provide; and by aligning each generated outcome with collision types, our design ensures controllability over collision types, which adversarial optimization approaches like STRIVE~\cite{rempe2022generating} and KING~\cite{hanselmann2022king} cannot guarantee.\\

{\noindent \bf Trajectory 
Prediction and Traffic Simulation.}
%In this section, we review existing  
% Since we formulate scenario generation as a structured trajectory prediction task conditioned on target collisions
Since our task involves generating future agent trajectories under collision constraints, it is most closely related to trajectory prediction, so we briefly review generative models for traffic simulation.
Early generative models such as Social-GAN \cite{gupta2018social} learn from past trajectories of each agent to predict diverse but plausible future motions.
% With past trajectories of each agent, Social-gan \cite{gupta2018social} predicts trajectories in the near future. 
TrafficSim \cite{suo2021trafficsim} and BITS \cite{xu2023bits} utilize map information as additional input to generate traffic flows.
However, these autoregressive predictors typically generate trajectories step by step, which often leads to error accumulation over longer horizons and makes it difficult to guarantee precise long-term outcomes such as specific collision types or TTA.
STRIVE~\cite{rempe2022generating} builds upon TrafficSim to generate safety-critical scenarios by optimizing perturbations in the latent space. 
More recently, diffusion-based predictors~\cite{gu2022stochastic, rempe2023trace} have 
demonstrated a stronger ability to capture the diverse distribution of multi-agent behaviors, alleviating the mode-collapse problem often observed in previous trajectory predictors.
CTG++~\cite{zhong2023language} introduces language-guided conditional generation via large language models (LLMs), where users provide prompts to influence future behavior. 
However, such natural-language conditions only guide agent behaviors at a coarse semantic level (e.g., intent or style), rather than enforcing precise spatial relations or collision timing.
% However, such conditions are often abstract or high-level, making it difficult to enforce precise collision types or time-to-accident control.

Our method instead adopts a back-to-front, coarse-to-fine generation process: rather than incrementally predicting each next step, which risks compounding errors and missing the desired outcome, we first predict the final collision pattern that encodes the collision type and TTA. The attacker’s full trajectory is then generated to realize this outcome while maintaining physical plausibility.\\
% Our method instead adopts a back-to-front, coarse-to-fine generation process: we begin by predicting the desired collision outcome, and then roll out full trajectories that realize this outcome. 
% This design allows structured scenario specification, enabling explicit control over collision type and urgency, an ability not afforded by existing trajectory predictors or traffic simulation models.

\begin{table}[t!]
\centering
\caption{\textbf{Comparison of existing collision-related datasets.} 
The table lists representative collision datasets with their primary task, the number of collision cases, and available attributes.}
% COLLIDE is designed for controllable scenario generation with the largest number of task-specific collision cases. Unlike prior datasets focusing on single-agent behavior or coarse attributes, COLLIDE provides structured labels for collision types and TTA.}

% \begin{tabular}{|l|l|l|l|}
% \begin{tabular}{ p{2.15cm} c p{1cm} p{1.73cm} }
% \begin{tabular}{ M{2cm} M{2.25cm} M{1cm} M{1.75cm} }
\begin{tabular}{@{}cc@{}c@{}c@{}}
% \hline
\toprule
Dataset & Task & \#Collision & Attributes \\ 
\midrule
YouTubeCrash \cite{kim2019crash} & collision prediction  & 122 & x \\
Street Accident \cite{chan2017anticipating} & collision prediction  & 678 & x \\
Collision \cite{herzig2019spatio} & collision detection   & 803 & x \\
VIENA \cite{aliakbarian2018viena} & behavior prediction  & 1200 & single behavior \\
CTA \cite{you2020traffic} & collision reasoning   & 1935 & collision cause \\
NIDB \cite{kataoka2018drive, suzuki2018anticipating} & collision prediction   & 4595 & topology \\
GTACrash \cite{kim2019crash} & collision prediction  & 7720 & x \\
RiskBench \cite{kung2024riskbench} & risk identification   & 1873 & single behavior \\
HazardVLM \cite{xiao2024hazardvlm} & hazard description   & 3860 & single behavior \\ 
\midrule %\hline
COLLIDE & scenario generation & \textbf{8586} & \textbf{collision type TTA} \\ 
% \hline
\bottomrule
\end{tabular}
\label{table:dataset}
\end{table}

% {\noindent \bf Datasets for Collision Scenario Generation.}
{\noindent \bf Collision Dataset.}
Existing real-world datasets such as YouTubeCrash~\cite{kim2019crash}, StreetAccidents~\cite{chan2017anticipating}, and VIENA~\cite{aliakbarian2018viena} have been widely used for training or evaluating models on accident anticipation or classification tasks. 
However, these datasets passively collect crash incidents from traffic camera footage or dashcams, offering biased coverage and lacking detailed semantic annotations such as collision types and time-to-accident (TTA).
Some simulation-based datasets, such as RiskBench~\cite{kung2024riskbench}, SafeBench~\cite{xu2022safebench}, and Target~\cite{deng2023target}, offer synthetic crash scenes for wider coverage of rare events, but they lack semantic labels for collision types, limited to per-vehicle behaviors rather than structured collision patterns.
To bridge this gap, we propose COLLIDE, a structured dataset derived from real-world logs via an automatic generation pipeline. 
It ensures balanced coverage across five NHTSA-defined collision types and TTA intervals, providing fine-grained control over scenario attributes.
As shown in Table \ref{table:dataset}, COLLIDE contains 8,586 task-specific collision cases, enabling scalable training and evaluation compared to prior datasets. Moreover, COLLIDE is the only dataset among existing works that supports scenario generation with explicit collision type control.

\section{Data Collection}
\label{sec:data_collection}

%======================Architecture=============================
%     {\noindent \bf 
% 1. Collision category \\
%     why we choose these 5 scenarios? \\
% 2. Data Collection \\
%     For both realistic and controllable demands, we need to build a dataset \\
% 3. Data augmentation \\
%     Furthermore, to include diversity and a wide distribution of TTC and low-level agent state (speed, yaw, position when collision), we need to augment data
% }
%======================Architecture=============================

\begin{figure*}[t!]
    \centering
    \includegraphics[width=\textwidth]{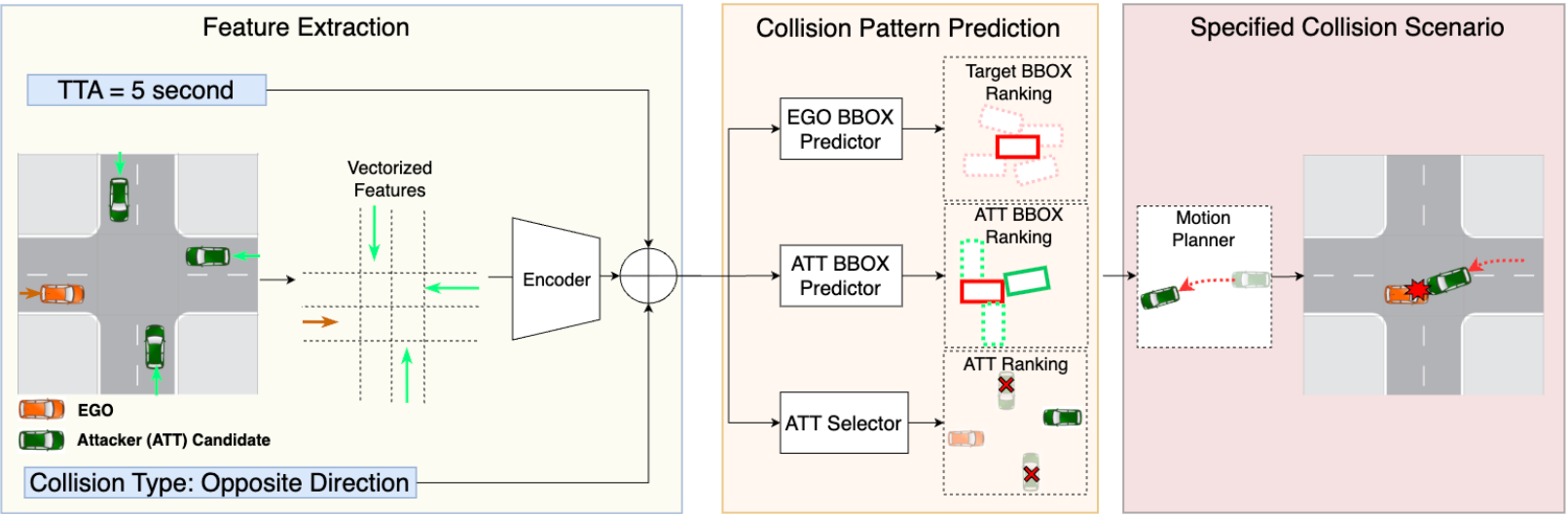}
    \caption{\textbf{The proposed architecture for controllable collision scenario generation}. Given structured scene context
    %a vectorized scene representation 
    and user-specified conditions (collision type and TTA), our model predicts the “collision pattern”, representing the relative configuration of ego and attacker vehicles (ATT) at the moment of collision. 
    %Three modules, Ego BBOX Predictor, Attacker BBOX Predictor, and Attacker Selector, are jointly trained to predict the collision pattern. The attacker trajectory is then generated by a quintic polynomial planner to realize this pattern.
    % \ychen{i think this sentence is incorrect. you do not generate plausible trajectory by training three modules.}
    %
    The predicted pattern then guides a motion planner to produce a feasible attacker trajectory.
    % This learning-based pipeline enables adaptive and diverse scenario synthesis beyond rule-based templates. 
 }
    \label{fig:Method}
\end{figure*}

This section introduces the definition of collision types, the data collection methodology, and scenario augmentation. 
%To train a learning-based model capable of generating controllable and diverse collision scenarios, we
We construct a dedicated dataset called \textbf{COLLIDE}, providing large-scale, balanced, and semantically grounded collision scenarios that are ideal for controllable scenario generation.

% \begin{figure*}[t!]
%     \centering
%     \includegraphics[width=\textwidth]{Figures/model_v4.pdf}
%     %\includegraphics[width=\linewidth]{Town10.pdf}
%     \caption{\textbf{The proposed architecture for controllable collision scenario generation}. Given structured scene context
%     %a vectorized scene representation 
%     and user-specified conditions (collision type and TTA), our model predicts the “collision pattern”, representing the relative configuration of ego and attacker vehicles (ATT) at the moment of collision. Three modules, Ego BBOX Predictor, Attacker BBOX Predictor, and Attacker Selector, are jointly trained to predict the collision pattern. The attacker trajectory is then generated by a quintic polynomial planner to realize this pattern.
%     % \ychen{i think this sentence is incorrect. you do not generate plausible trajectory by training three modules.}
%     %
%     This learning-based pipeline enables adaptive and diverse scenario synthesis beyond rule-based templates. 
%  }
%     \label{fig:Method}
% \end{figure*}

\subsection{Collision Types}
%To support systematic generation and evaluation of safety-critical scenarios, 
% To enable conditional collision scenario generation with fine-grained supervision,
We adopt five representative collision categories based on NHTSA \cite{najm2013description} statistics, which reflect 
%relative
societal impact across crash frequency, injury severity, and economic burden. These include: (1) Junction Crossing (JC), (2) Lane Change, (3) Opposite Direction, (4) Rear-End, and (5) Left-Turn Across Path (LTAP).
We use the relative heading angle at the collision point to assign categories: Rear-End ($\sim\!0^\circ$), Lane Change ($\sim\!20^\circ$), Opposite Direction ($\sim\!180^\circ$), Junction Crossing ($\sim\!90^\circ$), and LTAP ($\sim\!90^\circ$). A $10^\circ$ tolerance is applied to account for noise. 
Since both JC and LTAP involve a $90^\circ$ collision angle, we distinguish them by the relative heading when entering the intersection: JC maintains orthogonal paths throughout, while LTAP involves a left turn with vehicles approaching at a $180^\circ$ difference.

% Our formulation ensures diversity by constructing a balanced dataset with respect to collision types and time-to-accident (TTA), which provides a robust foundation for conditional scenario generation and planner evaluation. Fig. \ref{fig:circle} illustrates the uniform distribution achieved across the five collision types.

\subsection{Data Collection}

Given the scarcity of labeled collision scenarios in existing real-world datasets, we develop an automatic data collection pipeline that transforms \textbf{non-collision} scenes collected in the nuScenes dataset into synthetic \textbf{collision} scenarios. Each modified scene incorporates our self-defined collision pattern and preserves physical feasibility.

The original nuScenes scenarios span 20 seconds at 2Hz. We segment these into 4.5s to 9s clips, and at every 0.5-second interval, define a target collision point. The ego vehicle's bounding box serves as a reference to compute a collision target for an attacker. For example, in a rear-end scenario, the attacker must reach a predefined position behind the ego vehicle at the moment of collision.

A quintic polynomial planner is used to generate smooth attacker trajectories toward the target location. Candidate attackers are selected from other vehicles in the same scene that exhibit continuous motion. If a generated trajectory leads to infeasibility, such as collisions with other agents or off-road paths, the scenario is discarded. This ensures that the resulting dataset exhibits high control over both the collision type and TTA.

\subsection{Scenario Coverage} 

To ensure comprehensive coverage of safety-critical cases, our dataset incorporates multiple sources of variation.

\noindent\textbf{Collision geometry augmentation.} 
% Each of the five collision types is defined by geometric relationships between the ego and attacker vehicles, following NHTSA-inspired~\cite{najm2013description} criteria. 
% We use the relative heading angle at the collision point to assign categories: Rear-End ($\sim\!0^\circ$), Lane Change ($\sim\!20^\circ$), Opposite Direction ($\sim\!180^\circ$), Junction Crossing ($\sim\!90^\circ$), and LTAP ($\sim\!90^\circ$). A $10^\circ$ tolerance is applied to account for noise. 
% Since both JC and LTAP involve a $90^\circ$ collision angle, we distinguish them by the relative heading when entering the intersection: JC maintains orthogonal paths throughout, while LTAP involves a left turn with vehicles approaching at a $180^\circ$ difference. 
To enrich collision diversity, we sample collision angles uniformly within the defined margins to generate varied but semantically consistent scenarios.

\noindent\textbf{TTA augmentation.} 
% Building on the procedure in Sec. 3-B, we uniformly vary the target collision points along the ego trajectory, yielding scenarios that span TTAs from 4.5s to 9.0s, ensuring diversity in urgency levels the ego must respond to. 
% The uniform distribution over TTA intervals enhances coverage of low-TTA, high-risk cases that are critical for robust planner evaluation.
% For each ego trajectory, we first sample a target collision point along the future centerline of the ego vehicle. 
The time-to-accident (TTA) is defined as the temporal gap between the current time and the moment when the ego reaches the collision point. 
By uniformly sampling such collision points, we obtain TTAs distributed in the range of 4.5s to 9.0s, which balances between short-horizon, high-risk interactions and longer-horizon, low-urgency cases.

\noindent\textbf{Map diversity.} 
Our dataset covers all four cities in nuScenes, which provides diverse road structures and driving norms (e.g., left-hand vs. right-hand driving). This geographical diversity naturally enriches the collision scenarios with varied map topologies, ensuring broader coverage of intersection layouts, lane geometries, and driving conventions.

\begin{table*}[]
\caption{\textbf{Results of controllable collision scenario generation on COLLIDE.} 
% We compare with conditional variants (C-) of SGAN, STRIVE, and MID, where the desired collision type and TTA are provided as additional input. 
Our method achieves the highest collision rate and similarity across five collision types, outperforming conditional baselines.
% We evaluate collision rate and similarity across five collision types. 
% Similarity measures how well the generated scenario matches the specified collision type, reflecting alignment with user-defined conditions.
}
\label{table:method}
\centering
\resizebox{\textwidth}{!}{%
\begin{tabular}{llcccccc}
\toprule
\multicolumn{1}{c}{Method} &
  \multicolumn{1}{c}{Metric} &
  \multicolumn{1}{l}{Lane Change} &
  \multicolumn{1}{l}{Opposite Direction} &
  \multicolumn{1}{l}{Rear End} &
  \multicolumn{1}{l}{Junction Crossing} &
  \multicolumn{1}{l}{LTAP} &
  \multicolumn{1}{l}{Average} \\ 
  \midrule

% Collision rate row
C-SGAN \cite{gupta2018social}     & \multirow{4}{*}{\textbf{Collision Rate}}  & 20\% & 20\% & 18\% & 9\%  & 0\%  & 14\% \\
C-STRIVE \cite{rempe2022generating}  &                                   & 32\% & 21\% & 39\% & 16\% & 15\% & 25\% \\
C-MID   \cite{gu2022stochastic}   &                                   & 25\% & 10\% & 9\%  & 5\%  & 10\% & 11\% \\
\textbf{Ours}       &                                   & \textbf{73\%} & \textbf{77\%} & \textbf{84\%} & \textbf{81\%} & \textbf{90\%} & \textbf{81\%} \\ 

\midrule

% Similarity row
C-SGAN     & \multirow{4}{*}{\textbf{Similarity}} & 34\% & 33\%  & 68\% & 0\%  & 0\%  & 39\% \\
C-STRIVE   &                                      & 52\% & 75\%  & \textbf{93\%} & 52\% & 18\%  & 68\% \\ 
C-MID      &                                      & 42\% & 68\%  & 74\% & 33\% & 22\%  & 49\% \\
\textbf{Ours}       &                                      & \textbf{62}\% & \textbf{90\%}  & 87\% & \textbf{84\%} & \textbf{76\%}  & \textbf{81\%} \\
\bottomrule
\end{tabular}%
}
\end{table*}

\section{Method}

\subsection{Problem Formulation}

We aim to generate controllable collision scenarios conditioned on user-specified attributes, namely the desired collision type and time-to-accident (TTA). 
The input to the task consists of past trajectories of all traffic participants and the map topology. 
Each agent trajectory is represented as a sequence of positions and headings:
\(
s^{i}_{1:T_{\text{hist}}} = \{(x_i^t, y_i^t, \theta_i^t)\}_{t=1}^{T_{\text{hist}}},
\)
where $(x_i^t, y_i^t)$ denotes the 2D location of agent $i$ at time $t$, and $\theta_i^t$ is its heading angle. 
Let $\mathcal{M}$ denote the map topology. 
Given the historical trajectories of $N$ agents $\{s^i_{1:T_{\text{hist}}}\}_{i=1}^N$ and map $\mathcal{M}$, the goal is to generate an attacker trajectory
\(
s^{a}_{T_{\text{hist}}:t_{\text{TTA}}}
\)
that collides with the ego vehicle exactly at $t = t_{\text{TTA}}$. 
The generated trajectory is consistent with the user-specified collision type $\mathbf{C}_{\text{type}}$ and the urgency level determined by TTA. 
This formulation emphasizes two key challenges: (1) localizing where and how the collision occurs, and (2) realizing a kinematically feasible attacker trajectory that satisfies the specified conditions.

% \subsection{Problem Formulation}
% We aim to generate controllable collision scenarios conditioned on user-specified attributes. 
% Formally, given past trajectories of $N$ agents 
% \(
% s^{i}_{1:T_{\text{hist}}} = \{(x_i^t, y_i^t, \theta_i^t)\}_{t=1}^{T_{\text{hist}}}
% \)
% and the HD map $\mathcal{M}$, our goal is to produce an attacker trajectory 
% \(
% s^{a}_{T_{\text{hist}}:t_{\text{TTA}}}
% \)
% that collides with the ego vehicle at time $t=t_{\text{TTA}}$. 
% The generated scenario must match the user-specified collision type $\mathbf{C}_{\text{type}}$ and the temporal urgency defined by TTA. 
% This task requires localizing where and how the collision occurs, then realizing a kinematically feasible attacker trajectory consistent with this specification.

\subsection{Framework Overview}
As illustrated in Fig.~\ref{fig:Method}, 
% As illustrated in Fig.~\ref{fig:Method}, our framework adopts a back-to-front, coarse-to-fine strategy inspired by region proposal methods~\cite{ren2016faster} in object detection. 
rather than directly predicting the full attacker trajectory, we first predict a \textit{collision pattern}, a compact and interpretable representation of the final spatial configuration of the ego and attacker vehicles at the collision moment. 
Specifically, we define the collision pattern as a 4D vector
\(
\mathbf{p} = (\Delta x, \Delta y, \Delta \theta, \hat{a})
\)
where $(\Delta x, \Delta y)$ denotes the relative position of the attacker with respect to the ego vehicle, $\Delta \theta$ denotes their relative heading at impact, and $\hat{a}$ indicates the selected attacker.
The predicted pattern is then used as a target for trajectory realization by a quintic polynomial planner. 
% \plchen{The three modules are trained jointly in an end-to-end manner using supervision from ground-truth collision patterns in COLLIDE.}
Three trainable modules, Ego Position Prediction, Attacker Offset Prediction, and Attacker Selection, operate on the encoded scene and condition features to predict the collision pattern. 
% \textcolor{red}{you have the same content in the implementation details, where I feel is slightly better} \plchen{OK, I removed the repeated one.}

\subsection{Input Representation}
% \paragraph{Input: Scene and Condition Encoding.}
We adopt a VectorNet-based encoder~\cite{gao2020vectornet} to extract vectorized features from map topology and agent trajectories, forming the scene feature $\mathbf{F}_{\text{scene}}$. 
The user-specified condition, including collision type (one-hot vector $\mathbf{C}_{\text{type}}$) and normalized TTA value, is concatenated with $\mathbf{F}_{\text{scene}}$ to produce the final feature:
\(
\mathbf{F}_{\text{final}} = [\mathbf{F}_{\text{scene}}, \mathbf{C}_{\text{type}}, \text{TTA}].
\)
This final feature serves as the input to all subsequent modules.
% \paragraph{Pipeline.}
% The framework consists of three trainable modules for collision pattern prediction: Ego Position Prediction, Attacker Offset Prediction, and Attacker Selection. 
% Then followed by a quintic polynomial planner for trajectory realization. 
% All these modules generate controllable collision scenarios consistent with the given conditions.

\subsection{Collision Pattern Prediction}
\paragraph{Ego Position Prediction.}
We localize the anticipated collision point of the ego vehicle by framing it as a region proposal task. 
Candidate anchors $\{T_i\}$ are sampled along lane centerlines and classified as positive or negative depending on their proximity to the ground-truth collision point $x_{\text{ego, GT}}$ within radius $r$:
\[
\text{label}(T_i) = 
\begin{cases}
\text{positive}, & \|T_i - x_{\text{ego,GT}}\| \leq r, \\
\text{negative}, & \text{otherwise}.
\end{cases}
\]
Each candidate anchor $T_i$ is represented by a feature vector that includes its own spatial attributes (\( x,  y, \theta\)) together with concatenated local map 
%features aggregated from a neighborhood of fixed radius 
around $T_i$. 
For the top-ranked candidate, heading and offset regression produce the final ego bounding box $\mathbf{B}_{\text{ego}}(t_{\text{TTA}})$.
The local feature design allows the model to exploit richer local geometric information for more reliable ranking.

\paragraph{Attacker Offset Prediction.}
Given $\mathbf{B}_{\text{ego}}$ and $\mathbf{F}_{\text{final}}$, the model predicts a relative offset to determine the adversarial bounding box:
\[
\mathbf{B}_{\text{attacker}} = \mathbf{B}_{\text{ego}} + (\Delta x, \Delta y, \Delta \theta).
\]
This ensures controllable geometric relations consistent with the target collision type.

\paragraph{Attacker Selection.}
For each candidate agent $i$, its current state $\mathbf{P}_i$ is concatenated with $\mathbf{F}_{\text{final}}$ and scored via an MLP with softmax. 
The selected attacker is
\[
\hat{a} = \arg\max_i \, \text{MLP}([\mathbf{P}_i, \mathbf{F}_{\text{final}}]).
\]

\subsection{Trajectory Realization}
Given the predicted collision pattern, which specifies the relative spatial configuration of the ego and attacker at $t = T_{\text{TTA}}$, we employ a quintic polynomial planner to realize a smooth and kinematically feasible attacker trajectory. 
The collision pattern serves as the boundary condition for trajectory generation: the attacker must start from its observed history $s^a_{1:T_{\text{hist}}}$ and reach the configuration $\mathbf{B}_{\text{attacker}}$ exactly at the designated TTA. 
Formally, the trajectory is obtained as
\[
s^{\hat{a}}_{T_{\text{hist}}:T_{\text{TTA}}} = 
\mathcal{P}\!\left(s^a_{T_{\text{hist}}}, \mathbf{B}_{\text{attacker}}, T_{\text{TTA}}\right),
\]
where $\mathcal{P}$ denotes the polynomial planner. 
This construction guarantees physical feasibility while ensuring that the trajectory faithfully realizes the user-specified collision type and timing.

% \subsection{Trajectory Realization}
% Finally, given the attacker $\hat{a}$, the predicted target configuration $\mathbf{B}_{\text{attacker}}$, and TTA, we employ a quintic polynomial planner to generate a smooth, kinematically feasible trajectory:
% \[
% s^{\hat{a}}_{T_{\text{hist}}:t_{\text{TTA}}} = 
% \mathcal{P}\!\left(s^{\hat{a}}_{T_{\text{hist}}}, \mathbf{B}_{\text{attacker}}, \text{TTA}\right).
% \]
% This guarantees physical feasibility while ensuring that the resulting trajectory leads to the desired collision type and timing.

\section{Experimental Results}
\label{sec:result}

%======================Architecture=============================
%     {\noindent \bf 
% 1. Experiment setup: \\
%     a. Dataset: original nuScenes setting (time, Hz), our dataset setting \\
%     b. baselines: different setting between baselines (8 f input, 4 f output => open-loop) and ours \\
%     c. planner: replayed and rule-based planner from STRIVE \\
%     d. evaluation metrics: collision rate (interpolate, forward) and our new metric, similarity \\
% 2. Safety-critical scenario generation evaluation with baseline methods: (big table) \\
% 3. Ablation study on different setting (ours 3 setting) \\
% 4. Safety-critical scenario testing with different planners \\
% }
%======================Architecture=============================

\begin{figure*}[htb!]
    \centering
    \includegraphics[width=\textwidth]{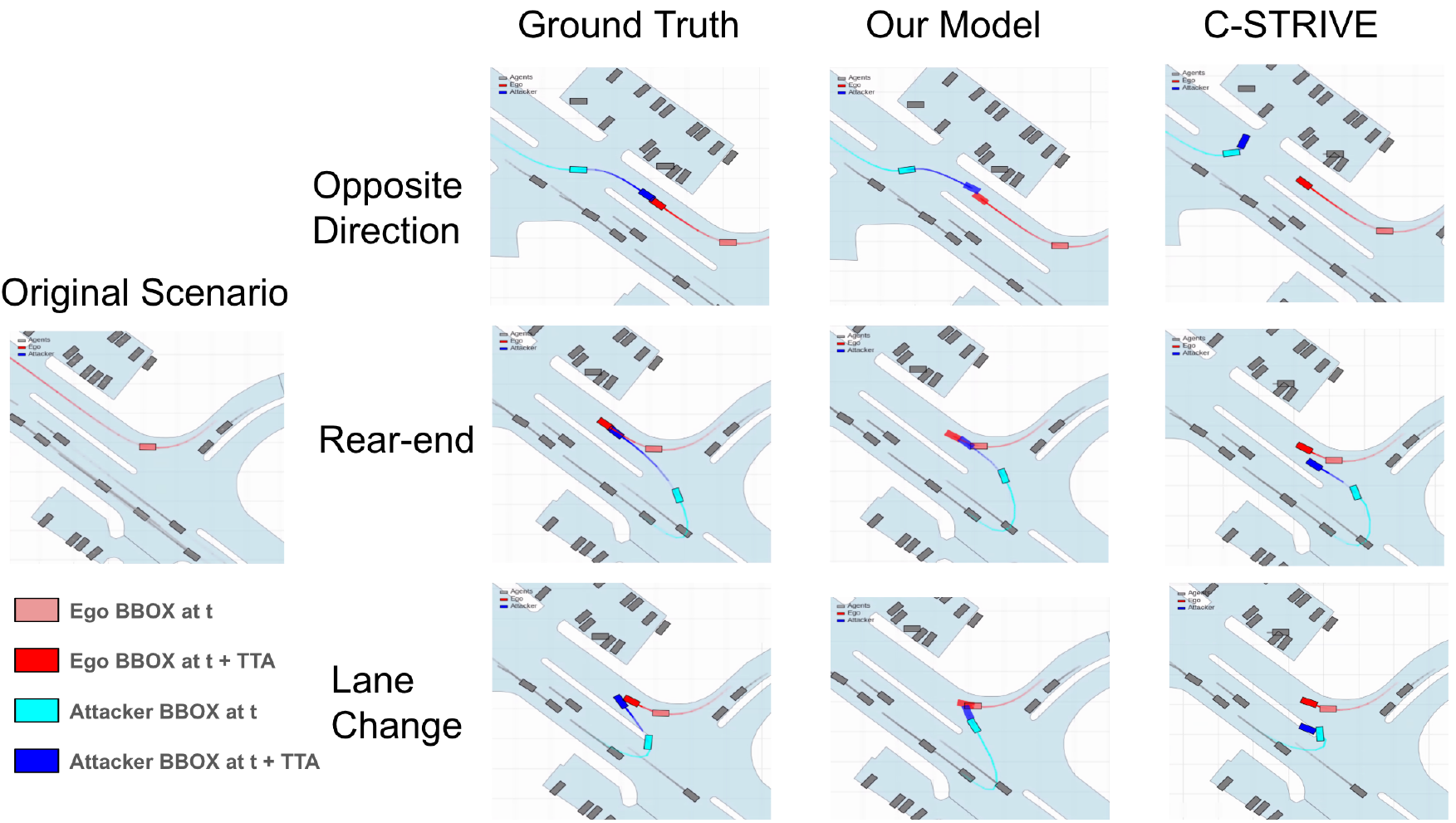}
    \caption{\textbf{Qualitative comparison of generated scenarios across different collision types.} We compare our dataset and model against C-STRIVE~\cite{rempe2022generating}, the strongest baseline. 
    Our models demonstrate strong controllability in generating user-specified collision types.
    % Our model better aligns with the desired collision patterns specified by our dataset (COLLIDE), whose ground truth guarantees collision type correctness.
    %
    % \ychen{i felt "Our Dataset" is not a proper wording. It should be "Ground Truth." }
    }

    \label{fig:experiment}
\end{figure*}

\subsection{Baselines and Evaluation Setup}
% To enable fair comparisons, we implement three conditional scenario generation baselines, C-SGAN, C-MID, and C-STRIVE, by adopting representative trajectory prediction models, Social-GAN~\cite{gupta2018social}, MID~\cite{gu2022stochastic}, and the traffic dynamic model of STRIVE~\cite{rempe2022generating}, respectively.
% We concatenate a one-hot vector of collision type and normalized time-to-accident (TTA) as the input of the decoder for the baselines.
We compare our method with three conditional scenario generation baselines, C-SGAN, C-MID, and C-STRIVE, implemented using Social-GAN~\cite{gupta2018social}, MID~\cite{gu2022stochastic}, and the traffic dynamic model of STRIVE~\cite{rempe2022generating}, respectively. For all baselines, a one-hot vector of collision type and the normalized time-to-accident (TTA) are concatenated as conditional inputs to the decoder. 
To ensure a fair comparison, baselines are extended to the same prediction horizon by applying open-loop autoregression: each iteration predicts 4 frames and appends them to the input sequence for the next step. Our method uses the same history input to directly predict a compact collision pattern, the terminal spatial configuration of the ego and attacker. A quintic polynomial planner then generates the full attacker trajectory to meet the predicted pattern at the desired TTA. This difference in generation strategy (pattern prediction + planner vs. autoregressive rollout) is explicitly accounted for in the evaluation by enforcing the same input horizon and final prediction horizon for all methods.

% This procedure approximates our design while requiring no changes to the baseline architectures.
% All approaches use 8 past frames as input. 
% Baselines predict 4 frames per iteration in an open-loop manner, appending their predictions to the input sequence for the next step. 
% This setting mimics our design without requiring architectural changes.

We do not compare with script-based or adversarial-based scenario 
generation methods, as they rely on pre-defined attacker scripts or 
agent-level adversarial optimization within simulation environments. 
These approaches are designed to explore possible crashes but are not 
directly applicable to offline driving logs such as COLLIDE, and they 
do not provide explicit controllability over collision type or TTA. 
Instead, we adapt conditional trajectory prediction models as baselines, 
since they naturally operate on logged data and can be extended to 
support generation conditioned on collision attributes.
% We choose trajectory prediction models as baselines instead of prior safety-critical scenario generation methods because our task is to learn from a curated set of labeled collision scenarios in a data-driven manner. 
% Most existing methods for safety-critical scenario generation, such as those based on scripted attacker behaviors or adversarial optimization, rely on agent-level control in simulation. 
% These are fundamentally incompatible with our goal of learning conditional trajectory distributions that can be directly applied to offline datasets.
% \textcolor{red}{unclear how the generated scenarios are evaluated}

\subsection{Evaluation Metrics}
To evaluate controllable scenario generation, we define two key metrics:
\begin{itemize}
\item \textbf{Collision Rate}: the percentage of modified attacker trajectories that result in a collision with the ego vehicle.
\item \textbf{Scenario Similarity}: the percentage of collisions in which the relative angle between ego and attacker matches the desired collision type, within a 10-degree tolerance.
\end{itemize}

\subsection{Implementation and Training Details}
Three modules, Ego Position Prediction, Attacker Offset Prediction, and Attacker Selection, are trained jointly in an end-to-end manner using supervision from ground-truth collision patterns extracted from COLLIDE. 
The full training loss is defined as a weighted sum of the ego position prediction loss, attacker offset regression loss, and attacker selection classification loss. 
We train the network for 1000 epochs using the Adam optimizer with a learning rate of $5 \times 10^{-5}$ and a batch size of 128. 
The overall inference latency is approximately 0.37 ms per scene on a single GPU, including both collision pattern prediction and polynomial trajectory realization.
% \textcolor{red}{"All three modules" => can be more specific, which three modules. Did the reviewers ask for latency? I feel it's ok to drop it.} 
% \plchen{According to Reviewer #14:"a brief discussion on the inference time per scenario would be beneficial to assess the scalability for building large-scale scenario libraries."}

\begin{table}[]

\caption{\textbf{Ablation study on ego position prediction module.} 
% We compare regression-based prediction, target point ranking, and region proposal in terms of displacement error, angle distance, collision rate, and similarity.
We evaluate different approaches for predicting the ego vehicle’s position at the collision moment, comparing regression-based prediction, target point ranking, and region proposal.
% The region proposal yields the best similarity and contributes most to the controllability observed in Table II.
%\ychen{font is too small}
}
\label{table:ablation}
\centering
\resizebox{\columnwidth}{!}{%
\begin{tabular}{lcccc}
\toprule
Setting              & \makecell{Displacement \\ Error (m)} & \makecell{Angle\\Distance ($^\circ$)} & \makecell{Collision \\ Rate} & Similarity \\ 
\midrule
Regression-based     &              3.95                 &            16                   &  69\%  & 67\%   \\
Target point ranking &              \textbf{2.15}                 &          11.4                     &  \textbf{80\%}  &  74\%  \\
Region proposal      &              3.1                 &          \textbf{10.18}                     &  77\%  &  \textbf{78\%} 
\\
\bottomrule
\end{tabular}%
}
\end{table}

\begin{table*}[t!]
\caption{
% We evaluate whether generated scenarios can consistently challenge a range of planners, including IDM~\cite{treiber2000congested}, rule-based from STRIVE~\cite{rempe2022generating}, and PDM~\cite{dauner2023parting}. 
\textbf{Evaluation of generated scenarios on autonomous vehicle planners.} We report collision rates of C-STRIVE and our method across five collision types, tested on different planners.
% We evaluate the impact of the generated scenarios by testing them on AV planners.
Higher collision rates indicate stronger effectiveness in exposing planner failure cases. 
% \hank{what is the metric?}
\label{table:planner evaluation}
}
\centering
\resizebox{\textwidth}{!}{%
\begin{tabular}{llcccccc}
\toprule
\multicolumn{1}{c}{Planner} &
  \multicolumn{1}{c}{Algorithms} &
  \multicolumn{1}{l}{lane change} &
  \multicolumn{1}{l}{opposite direction} &
  \multicolumn{1}{l}{rear end} &
  \multicolumn{1}{l}{junction crossing} &
  \multicolumn{1}{l}{LTAP} &
  \multicolumn{1}{l}{Average} \\ 
  \midrule
\multirow{2}{*} {IDM~\cite{treiber2000congested}}    
 &    C-STRIVE                               & \textbf{17\%} & 11\% & 58\% & 9\% & 4\% & 21\% \\
 &    Ours                               & 16\% & \textbf{30\%} & \textbf{65\%} & \textbf{20\%} & \textbf{6\%} & \textbf{29\%} \\ \midrule

\multirow{2}{*} {Rule-based~\cite{rempe2022generating}}     
 &    C-STRIVE                               & 23\% & 24\% & \textbf{40\%} & 13\% & \textbf{23\%} & 25\%
  \\
&  Ours                                    & \textbf{31\%} & \textbf{52\%}  & 18\% & \textbf{28\%} & 22\%  & \textbf{30\%} \\ 
\midrule

\multirow{2}{*} {PDM~\cite{dauner2023parting}}     
 &    C-STRIVE                               & 12\% & 12\% & 67\% & 11\% & 11\% & 24\%
  \\
&  Ours                                    & \textbf{19\%} & \textbf{37\%}  & \textbf{76\%} & \textbf{18\%} & \textbf{39\%}  & \textbf{39\%} \\ 
\bottomrule
\end{tabular}%
}
\end{table*}

\subsection{Results of Conditional Collision Scenario Generation}
Tab. \ref{table:method} shows that our method consistently achieves higher collision rates and scenario similarity across all five collision types compared to conditional baselines. 
The generated scenarios not only cause more collisions but also better match the intended collision patterns specified by time-to-accident (TTA) and collision type.
Fig.~\ref{fig:experiment} highlights common failure cases of the baselines. 
For instance, in the Opposite Direction example, C-STRIVE~\cite{rempe2022generating} predicts U-turn behaviors rather than intrusions into the ego lane, resulting in high deviation from the ground-truth trajectory. 
Similarly, in Rear-End scenarios, C-STRIVE~\cite{rempe2022generating} fails to capture overtaking attackers that decelerate sharply in front of the ego and instead defaults to benign forward-driving behaviors. 
In contrast, C-STRIVE~\cite{rempe2022generating} performs well in the common scenario where both vehicles are driving in the same direction.
This limitation arises because C-STRIVE lacks fine-grained collision type guidance and therefore tends to overfit to the dominant prevalent, and safer behaviors in the dataset.
By introducing collision patterns as explicit intermediate targets, our method overcomes this bias and reliably generates hazardous yet underrepresented events consistent with the specified collision type and TTA.

% These failure cases highlight the bias of conditional trajectory predictors toward frequent, benign maneuvers. 
% By introducing collision patterns as compact
% and interpretable intermediate targets, our method circumvents this bias: 
% The model is explicitly guided to realize rare but safety-critical outcomes 
% while still respecting the specified collision type and TTA. 
% This design ensures that hazardous but underrepresented events are systematically captured, rather than overlooked by data-driven frequency bias.

% Fig.~\ref{fig:experiment} shows typical failure cases of C-STRIVE~\cite{rempe2022generating}: it often predicts benign maneuvers (e.g., U-turns or straight driving) instead of rare but safety-critical intrusions such as opposite-direction or sharp rear-end collisions. 
% This bias stems from the lack of fine-grained collision-type guidance, causing the model to overfit to prevalent safe behaviors. 
% By introducing collision patterns as explicit intermediate targets, our method overcomes this bias and reliably generates hazardous yet underrepresented events consistent with the specified collision type and TTA.

%These examples illustrate the bias of trajectory prediction models toward common, safe maneuvers and their limited capacity to capture rare but hazardous behaviors.

%In contrast, our method better captures these aggressive patterns while adhering to the target TTA and collision type, enabling more diverse and controllable scenario generation.

\subsection{Ablation Study on Ego Bounding Box Prediction}
% Tab. \ref{table:ablation} evaluates the impact of different strategies on collision pattern prediction.
% Since the collision pattern prediction can be decomposed into predicting the endpoint of the future trajectory for two vehicles, one intuitive approach is to use a regression module to predict the relative distance from the starting position to the endpoint for both the ego and attacker, with the regression-based setting serving as our baseline.

% The target point ranking method follows the approach used in TNT \cite{zhao2021tnt}, a goal-oriented trajectory prediction method with a VectorNet~\cite{gao2020vectornet} backbone. 
% It first selects target points along the centerline of the topology at regular intervals, and then uses vectorized features to rank these target points for selecting the most likely position. 
% Finally, it refines the final prediction via an offset regression in a coarse-to-fine process. 
% This setting reduces displacement and angular errors but still lacks fine-grained topological alignment.

% Our final approach is a region proposal-based design that attaches local subgraph features to each anchor and classifies whether a position is likely to be a collision target. 
% This local view emphasizes lane-conforming orientation, e.g. predicting vehicle headings aligned with the road structure, which significantly improves angular precision and achieves the highest scenario similarity.

We investigate three strategies for localizing the ego collision endpoint, a key step for shaping the collision pattern. 
The regression-based approach directly predicts offsets in $(x,y,\theta)$, but often produces imprecise goals. 
Target point ranking~\cite{zhao2021tnt} improves positional accuracy by selecting among candidate anchors along centerlines, yet may still misalign with the lane topology. 
Our region proposal design further attaches local subgraph features to anchors, emphasizing lane-conforming orientations. 

Tab.~\ref{table:ablation} shows that regression yields the largest errors, ranking reduces displacement but lacks angular precision, while region proposals achieve the best similarity to ground-truth patterns. 
Since our task prioritizes semantic consistency of collision patterns over raw collision frequency, we adopt the region proposal variant in the final framework.

% From the experiments, we infer that the global features provide better overall scene understanding, allowing for more accurate collision position predictions, leading to lower displacement errors and higher collision rates. However, the region-proposal-based method, which utilizes local features, makes better use of topology information around the target points. For example, the vehicle’s heading should be parallel to the lane, which improves angle prediction. This approach achieves the highest scenario similarity, which is crucial for our conditional scenario generation task.

While the target ranking strategy better predicts whether a collision will occur, the region proposal method more faithfully captures how the collision happens. 
Since our goal is to generate controllable scenarios that match specific types, we prioritize semantic consistency and pattern fidelity over mere collision frequency. 
Therefore, we adopt the region proposal variant in our final model.

% \subsection{Planner Evaluation}

% To evaluate the practical applicability of our method in AV safety assessment, we replace the behavior of the ego vehicle, originally recorded in the dataset, with different planning algorithms. As shown in Tab \ref{table:planner evaluation}, the collision scenarios generated by our method consistently lead to higher collision rates across various scenario types compared to those generated by C-STRIVE, indicating that our scenarios remain more robust than baselines.

% In our evaluation, the IDM planner treats the ground-truth attacker as the leading vehicle to monitor, resulting in a relatively low collision rate due to its conservative car-following behavior. For comparison, we also include the rule-based planner proposed by STRIVE and the PDM planner, the winner of the 2023 nuPlan Planning Challenge. Both are evaluated with default parameters. The key difference is that the STRIVE planner follows the ego vehicle’s original trajectory, whereas the PDM planner uses the lane centerline as a reference path and performs online prediction to identify potential leading vehicles. This increased uncertainty in the PDM’s perception of front vehicles results in higher collision rates.

\subsection{Planner Evaluation under Safety-Critical Scenarios}

To assess the practical impact of our generated scenarios, we evaluate three planners: IDM~\cite{treiber2000congested}, the STRIVE~\cite{rempe2022generating} rule-based planner, and the PDM~\cite{dauner2023parting} planner from the nuPlan 2023 planning challenge, under scenarios generated by our method and C-STRIVE. As shown in Tab. \ref{table:planner evaluation}, our scenarios consistently lead to higher collision rates across all scenario types and planners.

Both the IDM and STRIVE rule-based planners rely on the original ego trajectory as a reference path. The IDM planner treats the attacker as a lead vehicle and follows a conservative car-following policy, often resulting in early braking and thus lower collision rates. In contrast, the STRIVE rule-based planner applies a simple brake–accelerate rule to slightly adjust the replayed speed, producing behavior more similar to the original collision scenario.
The PDM planner 
% adopts a more complex planning pipeline: it 
uses the lane centerline as a reference path, generates candidate maneuvers in both lateral and longitudinal directions, and evaluates them using an internal scoring function. To assess potential collisions, it simulates the future motion of surrounding vehicles under the assumption that each continues with its current velocity rather than relying on learned trajectory predictors.
This uncertainty in reasoning about the front agent leads to a higher collision rate.

The results confirm that our method produces more adversarial scenarios, capable of consistently triggering planner failures even under motion planning setups.

\subsection{Improving Planner with Generated Scenarios}

\textbf{Planner optimization process.} To demonstrate the utility of our generated collision scenarios, we conduct parameter optimization of the PDM planner~\cite{dauner2023parting} via grid search under both C-STRIVE-generated scenarios and those from our method (Tab.~\ref{table:planner improvement}). 
The planner constructs candidate trajectories by laterally shifting the centerline reference path of the ego vehicle to simulate evasive maneuvers and combining each shifted path with different velocity scales to simulate varying levels of deceleration. 
Each candidate trajectory is evaluated with the IDM policy, and the best one is selected according to a scoring function. 
We tune both IDM parameters (e.g., minimum desired distance, maximum acceleration) and the scoring weights (e.g., progress and timing weights).

\textbf{Results.} C-STRIVE tends to yield overly aggressive attacker behaviors, often leading to collisions earlier than the specified TTA. 
This drives the PDM planner to favor more frequent lane changes, which is advantageous in scenarios such as \textit{Opposite Direction} (OD) and \textit{Rear-End (RE)}, where collision avoidance is only feasible through lateral maneuvers. 
These results highlight that different scenario generation methods expose different planner failure modes, underscoring the need for a broader and more controllable set of testing scenarios to comprehensively assess planner robustness and generalization. Qualitative visualizations of planner behavior after fine-tuning with the generated scenarios are provided in the \href{https://plchen86157.github.io/conditional_scenario_generation/}{project webpage}.

\begin{table}[]
\caption{
%We select PDM~\cite{dauner2023parting}, the winning planner of the nuPlan 2023 Planning Challenge, a４s the target for parameter optimization. 
\textbf{Parameter tuning results on planners.}
We tune the policy parameters of PDM to evaluate collision rate using scenarios generated by different methods.}
% The table reports the collision rate of PDM across five collision types.
% : Lane Change (LC), Opposite Direction (OD), Rear End (RE), Junction Crossing (JC), and LTAP.}

% Our generated scenarios effectively guide parameter tuning and lead to improved planner safety.}
\label{table:planner improvement}
\centering
\resizebox{\columnwidth}{!}{%
\begin{tabular}{llllll}
\toprule
Parameter              & LC & OD & RE & JC & LTAP\\ 
\midrule
Default     &     \textbf{8.1}         \%                 &     43.4  \%                        &  72.6\%  &  14.9\% & 27.8\%  \\ 
C-STRIVE's       &   8.7         \%                 &     \textbf{39.4}        \%                  &  \textbf{70.3\%}  &  12.9\% & 20.9\% \\
Ours      &      8.7         \%                 &     40.0     \%                     &  70.4\%  &  \textbf{12.5\%} & \textbf{20.8\%} 
\\
\bottomrule
\end{tabular}%
}
\end{table}

%{\hspace*{\fill}Route Comp (\%) ↑\hspace*{\fill}}

\section{Conclusion and Future Works}
% In this unexplored task, we collect a self-designed dataset to correctly reconstruct a desired collision scenario for AVs' robustness. We furthermore prove that our conditional collision scenario algorithm, inspired by region proposal, dramatically outperforms other baselines on both collision rate and similarity. We hope to incorporate large language models (LLMs) in the future to make the condition inputs more flexible and user-friendly. In the subsequent experiments, we show the effectiveness of our generated scenarios on different planners and demonstrate that our model can improve the rule-based planner’s ability. As a potential direction for future work, our current trajectory-level scenario generation could be extended to the image level, enabling direct testing of end-to-end models that require visual input.

We present controllable collision scenario generation as a new task for systematic AV safety evaluation. To this end, we build COLLIDE, the first dataset with balanced coverage of collision types and time-to-accident (TTA). We further introduce the concept of Collision Pattern and design a region-proposal-inspired framework that enables fine-grained control over collision outcomes. Experiments show that our method outperforms strong baselines in controllability and plausibility, while also revealing planner limitations and improving their robustness.

\section*{Acknowledgement}
This work is sponsored in part by the National Science and Technology Council under grants 113-2628-E-A49-022- and 114-2628-E-A49-007-, MobileDrive, the Higher Education Sprout Project of National Yang Ming Chiao Tung University, the Ministry of Education, and the Yushan Fellow Program Administrative Support Grant.

% --- Bibliography ---
\bibliographystyle{IEEEtran}
% \bibliography{egbib}

\begin{thebibliography}{10}
\providecommand{\url}[1]{#1}
\csname url@samestyle\endcsname
\providecommand{\newblock}{\relax}
\providecommand{\bibinfo}[2]{#2}
\providecommand{\BIBentrySTDinterwordspacing}{\spaceskip=0pt\relax}
\providecommand{\BIBentryALTinterwordstretchfactor}{4}
\providecommand{\BIBentryALTinterwordspacing}{\spaceskip=\fontdimen2\font plus
\BIBentryALTinterwordstretchfactor\fontdimen3\font minus \fontdimen4\font\relax}
\providecommand{\BIBforeignlanguage}[2]{{%
\expandafter\ifx\csname l@#1\endcsname\relax
\typeout{** WARNING: IEEEtran.bst: No hyphenation pattern has been}%
\typeout{** loaded for the language `#1'. Using the pattern for}%
\typeout{** the default language instead.}%
\else
\language=\csname l@#1\endcsname
\fi
#2}}
\providecommand{\BIBdecl}{\relax}
\BIBdecl

\bibitem{ding2023survey}
W.~Ding, C.~Xu, M.~Arief, H.~Lin, B.~Li, and D.~Zhao, ``A survey on safety-critical driving scenario generation - a methodological perspective,'' \emph{IEEE Transactions on Intelligent Transportation Systems}, vol.~24, no.~7, pp. 6971--6988, 2023.

\bibitem{schutt20231001}
B.~Sch{\"u}tt, J.~Ransiek, T.~Braun, and E.~Sax, ``1001 ways of scenario generation for testing of self-driving cars: A survey,'' in \emph{2023 IEEE Intelligent Vehicles Symposium (IV)}.\hskip 1em plus 0.5em minus 0.4em\relax IEEE, 2023, pp. 1--8.

\bibitem{gao2025foundation}
Y.~Gao, M.~Piccinini, Y.~Zhang, D.~Wang, K.~Moller, R.~Brusnicki, B.~Zarrouki, A.~Gambi, J.~F. Totz, K.~Storms \emph{et~al.}, ``Foundation models in autonomous driving: A survey on scenario generation and scenario analysis,'' \emph{arXiv preprint arXiv:2506.11526}, 2025.

\bibitem{wang2024survey}
Z.~Wang, J.~Ma, and E.~M. Lai, ``A survey of scenario generation for automated vehicle testing and validation,'' \emph{Future Internet}, vol.~16, no.~12, p. 480, 2024.

\bibitem{Kung_2024_CVPR}
C.-H. Kung, S.-W. Lu, Y.-H. Tsai, and Y.-T. Chen, ``Action-slot: Visual action-centric representations for multi-label atomic activity recognition in traffic scenes,'' in \emph{Proceedings of the IEEE/CVF Conference on Computer Vision and Pattern Recognition (CVPR)}, June 2024, pp. 18\,451--18\,461.

\bibitem{liu2024curse}
H.~X. Liu and S.~Feng, ``Curse of rarity for autonomous vehicles,'' \emph{nature communications}, vol.~15, no.~1, p. 4808, 2024.

\bibitem{rempe2022generating}
D.~Rempe, J.~Philion, L.~J. Guibas, S.~Fidler, and O.~Litany, ``Generating useful accident-prone driving scenarios via a learned traffic prior,'' in \emph{Proceedings of the IEEE/CVF Conference on Computer Vision and Pattern Recognition}, 2022, pp. 17\,305--17\,315.

\bibitem{hanselmann2022king}
N.~Hanselmann, K.~Renz, K.~Chitta, A.~Bhattacharyya, and A.~Geiger, ``King: Generating safety-critical driving scenarios for robust imitation via kinematics gradients,'' in \emph{European Conference on Computer Vision}.\hskip 1em plus 0.5em minus 0.4em\relax Springer, 2022, pp. 335--352.

\bibitem{rempe2023trace}
D.~Rempe, Z.~Luo, X.~Bin~Peng, Y.~Yuan, K.~Kitani, K.~Kreis, S.~Fidler, and O.~Litany, ``Trace and pace: Controllable pedestrian animation via guided trajectory diffusion,'' in \emph{Proceedings of the IEEE/CVF Conference on Computer Vision and Pattern Recognition}, 2023, pp. 13\,756--13\,766.

\bibitem{gu2022stochastic}
T.~Gu, G.~Chen, J.~Li, C.~Lin, Y.~Rao, J.~Zhou, and J.~Lu, ``Stochastic trajectory prediction via motion indeterminacy diffusion,'' in \emph{Proceedings of the IEEE/CVF conference on computer vision and pattern recognition}, 2022, pp. 17\,113--17\,122.

\bibitem{chan2017anticipating}
F.-H. Chan, Y.-T. Chen, Y.~Xiang, and M.~Sun, ``Anticipating accidents in dashcam videos,'' in \emph{Computer Vision--ACCV 2016: 13th Asian Conference on Computer Vision, Taipei, Taiwan, November 20-24, 2016, Revised Selected Papers, Part IV 13}.\hskip 1em plus 0.5em minus 0.4em\relax Springer, 2017, pp. 136--153.

\bibitem{kataoka2018drive}
H.~Kataoka, T.~Suzuki, S.~Oikawa, Y.~Matsui, and Y.~Satoh, ``Drive video analysis for the detection of traffic near-miss incidents,'' in \emph{2018 IEEE International Conference on robotics and automation (ICRA)}.\hskip 1em plus 0.5em minus 0.4em\relax IEEE, 2018, pp. 3421--3428.

\bibitem{suzuki2018anticipating}
T.~Suzuki, H.~Kataoka, Y.~Aoki, and Y.~Satoh, ``Anticipating traffic accidents with adaptive loss and large-scale incident db,'' in \emph{Proceedings of the IEEE conference on computer vision and pattern recognition}, 2018, pp. 3521--3529.

\bibitem{kim2019crash}
H.~Kim, K.~Lee, G.~Hwang, and C.~Suh, ``Crash to not crash: Learn to identify dangerous vehicles using a simulator,'' in \emph{Proceedings of the AAAI Conference on Artificial Intelligence}, vol.~33, no.~01, 2019, pp. 978--985.

\bibitem{you2020traffic}
T.~You and B.~Han, ``Traffic accident benchmark for causality recognition,'' in \emph{Computer Vision--ECCV 2020: 16th European Conference, Glasgow, UK, August 23--28, 2020, Proceedings, Part VII 16}.\hskip 1em plus 0.5em minus 0.4em\relax Springer, 2020, pp. 540--556.

\bibitem{xiao2024hazardvlm}
D.~Xiao, M.~Dianati, P.~Jennings, and R.~Woodman, ``Hazardvlm: A video language model for real-time hazard description in automated driving systems,'' \emph{IEEE Transactions on Intelligent Vehicles}, 2024.

\bibitem{kung2024riskbench}
C.-H. Kung, C.-C. Yang, P.-Y. Pao, S.-W. Lu, P.-L. Chen, H.-C. Lu, and Y.-T. Chen, ``Riskbench: A scenario-based benchmark for risk identification,'' in \emph{2024 IEEE International Conference on Robotics and Automation (ICRA)}.\hskip 1em plus 0.5em minus 0.4em\relax IEEE, 2024, pp. 14\,800--14\,807.

\bibitem{najm2013description}
W.~G. Najm, R.~Ranganathan, G.~Srinivasan, J.~D. Smith, S.~Toma, E.~D. Swanson, A.~Burgett \emph{et~al.}, ``Description of light-vehicle pre-crash scenarios for safety applications based on vehicle-to-vehicle communications,'' United States. Department of Transportation. National Highway Traffic Safety..., Tech. Rep., 2013.

\bibitem{ren2016faster}
S.~Ren, K.~He, R.~Girshick, and J.~Sun, ``Faster r-cnn: Towards real-time object detection with region proposal networks,'' \emph{IEEE transactions on pattern analysis and machine intelligence}, vol.~39, no.~6, pp. 1137--1149, 2016.

\bibitem{gupta2018social}
A.~Gupta, J.~Johnson, L.~Fei-Fei, S.~Savarese, and A.~Alahi, ``Social gan: Socially acceptable trajectories with generative adversarial networks,'' in \emph{Proceedings of the IEEE conference on computer vision and pattern recognition}, 2018, pp. 2255--2264.

\bibitem{treiber2000congested}
M.~Treiber, A.~Hennecke, and D.~Helbing, ``Congested traffic states in empirical observations and microscopic simulations,'' \emph{Physical review E}, vol.~62, no.~2, p. 1805, 2000.

\bibitem{dauner2023parting}
D.~Dauner, M.~Hallgarten, A.~Geiger, and K.~Chitta, ``Parting with misconceptions about learning-based vehicle motion planning,'' in \emph{Conference on Robot Learning}.\hskip 1em plus 0.5em minus 0.4em\relax PMLR, 2023, pp. 1268--1281.

\bibitem{knies2020data}
C.~Knies and F.~Diermeyer, ``Data-driven test scenario generation for cooperative maneuver planning on highways,'' \emph{Applied Sciences}, vol.~10, no.~22, p. 8154, 2020.

\bibitem{scanlon2021waymo}
J.~M. Scanlon, K.~D. Kusano, T.~Daniel, C.~Alderson, A.~Ogle, and T.~Victor, ``Waymo simulated driving behavior in reconstructed fatal crashes within an autonomous vehicle operating domain,'' \emph{Accident Analysis \& Prevention}, vol. 163, p. 106454, 2021.

\bibitem{ding2018new}
W.~Ding, W.~Wang, and D.~Zhao, ``A new multi-vehicle trajectory generator to simulate vehicle-to-vehicle encounters,'' \emph{arXiv preprint arXiv:1809.05680}, 2018.

\bibitem{krajewski2018highd}
R.~Krajewski, J.~Bock, L.~Kloeker, and L.~Eckstein, ``The highd dataset: A drone dataset of naturalistic vehicle trajectories on german highways for validation of highly automated driving systems,'' in \emph{2018 21st international conference on intelligent transportation systems (ITSC)}.\hskip 1em plus 0.5em minus 0.4em\relax IEEE, 2018, pp. 2118--2125.

\bibitem{wang2021advsim}
J.~Wang, A.~Pun, J.~Tu, S.~Manivasagam, A.~Sadat, S.~Casas, M.~Ren, and R.~Urtasun, ``Advsim: Generating safety-critical scenarios for self-driving vehicles,'' in \emph{Proceedings of the IEEE/CVF Conference on Computer Vision and Pattern Recognition}, 2021, pp. 9909--9918.

\bibitem{suo2021trafficsim}
S.~Suo, S.~Regalado, S.~Casas, and R.~Urtasun, ``Trafficsim: Learning to simulate realistic multi-agent behaviors,'' in \emph{Proceedings of the IEEE/CVF Conference on Computer Vision and Pattern Recognition}, 2021, pp. 10\,400--10\,409.

\bibitem{ding2023causalaf}
W.~Ding, H.~Lin, B.~Li, and D.~Zhao, ``Causalaf: Causal autoregressive flow for safety-critical driving scenario generation,'' in \emph{Conference on robot learning}.\hskip 1em plus 0.5em minus 0.4em\relax PMLR, 2023, pp. 812--823.

\bibitem{huang2024cadre}
P.~Huang, W.~Ding, B.~Stoler, J.~Francis, B.~Chen, and D.~Zhao, ``Cadre: Controllable and diverse generation of safety-critical driving scenarios using real-world trajectories,'' \emph{arXiv preprint arXiv:2403.13208}, 2024.

\bibitem{xu2022safebench}
C.~Xu, W.~Ding, W.~Lyu, Z.~Liu, S.~Wang, Y.~He, H.~Hu, D.~Zhao, and B.~Li, ``Safebench: A benchmarking platform for safety evaluation of autonomous vehicles,'' \emph{Advances in Neural Information Processing Systems}, vol.~35, pp. 25\,667--25\,682, 2022.

\bibitem{xu2023bits}
D.~Xu, Y.~Chen, B.~Ivanovic, and M.~Pavone, ``Bits: Bi-level imitation for traffic simulation,'' in \emph{2023 IEEE International Conference on Robotics and Automation (ICRA)}.\hskip 1em plus 0.5em minus 0.4em\relax IEEE, 2023, pp. 2929--2936.

\bibitem{zhong2023language}
Z.~Zhong, D.~Rempe, Y.~Chen, B.~Ivanovic, Y.~Cao, D.~Xu, M.~Pavone, and B.~Ray, ``Language-guided traffic simulation via scene-level diffusion,'' in \emph{Conference on Robot Learning}.\hskip 1em plus 0.5em minus 0.4em\relax PMLR, 2023, pp. 144--177.

\bibitem{herzig2019spatio}
R.~Herzig, E.~Levi, H.~Xu, H.~Gao, E.~Brosh, X.~Wang, A.~Globerson, and T.~Darrell, ``Spatio-temporal action graph networks,'' in \emph{Proceedings of the IEEE/CVF international conference on computer vision workshops}, 2019, pp. 0--0.

\bibitem{aliakbarian2018viena}
M.~S. Aliakbarian, F.~S. Saleh, M.~Salzmann, B.~Fernando, L.~Petersson, and L.~Andersson, ``Viena: A driving anticipation dataset,'' in \emph{Asian Conference on Computer Vision}.\hskip 1em plus 0.5em minus 0.4em\relax Springer, 2018, pp. 449--466.

\bibitem{deng2023target}
\BIBentryALTinterwordspacing
Y.~Deng, J.~Yao, Z.~Tu, X.~Zheng, M.~Zhang, and T.~Zhang, ``Target: Automated scenario generation from traffic rules for testing autonomous vehicles,'' \emph{arXiv preprint arXiv:2305.06018}, 2023. [Online]. Available: \url{https://arxiv.org/abs/2305.06018}
\BIBentrySTDinterwordspacing

\bibitem{gao2020vectornet}
J.~Gao, C.~Sun, H.~Zhao, Y.~Shen, D.~Anguelov, C.~Li, and C.~Schmid, ``Vectornet: Encoding hd maps and agent dynamics from vectorized representation,'' in \emph{Proceedings of the IEEE/CVF conference on computer vision and pattern recognition}, 2020, pp. 11\,525--11\,533.

\bibitem{zhao2021tnt}
H.~Zhao, J.~Gao, T.~Lan, C.~Sun, B.~Sapp, B.~Varadarajan, Y.~Shen, Y.~Shen, Y.~Chai, C.~Schmid \emph{et~al.}, ``Tnt: Target-driven trajectory prediction,'' in \emph{Conference on Robot Learning}.\hskip 1em plus 0.5em minus 0.4em\relax PMLR, 2021, pp. 895--904.

\end{thebibliography}

\end{document}